\documentclass[a4paper,11pt]{article}
\usepackage[english]{babel}
\usepackage[utf8]{inputenc}
\usepackage{csquotes}
\usepackage[table,dvipsnames]{xcolor}
\usepackage{graphicx}
\usepackage{url}
\usepackage{amsmath}
\usepackage{bm}
\usepackage{geometry}
\usepackage{coling2020}
\usepackage{times}
\usepackage{latexsym}
\usepackage{microtype}
\usepackage[normalem]{ulem}

\title{UoB at SemEval-2020 Task 1: Automatic Identification of Novel Word Senses}

\author{Eleri Sarsfield\\
  School of Computer Science \\
  University of Birmingham\\
  United Kingdom \\
  {\small \tt elerisarsfield@gmail.com} \\\And
Harish Tayyar Madabushi \\
  School of Computer Science \\
  University of Birmingham\\
  United Kingdom \\
  {\small \tt H.TayyarMadabushi.1@bham.ac.uk} \\
}
\date{}
\colingfinalcopy
\begin{document}
\maketitle
\begin{abstract}
Much as the social landscape in which languages are spoken shifts, language too evolves to suit the needs of its users. Lexical semantic change analysis is a burgeoning field of semantic analysis which aims to trace changes in the meanings of words over time.  This paper presents an approach to lexical semantic change detection based on Bayesian word sense induction suitable for novel word sense identification. This approach is used for a submission to SemEval-2020 Task 1, which shows the approach to be capable of the SemEval task. The same approach is also applied to a corpus gleaned from 15 years of Twitter data, the results of which are then used to identify words which may be instances of slang.
\end{abstract}

\section{Introduction}
\label{sec:intro}
\blfootnote{
    \hspace{-0.65cm}  
    Accepted for publication at SemEval 2020 \\
    \hspace{-0.65cm}  
     This work is licensed under a Creative Commons 
     Attribution 4.0 International Licence.
     Licence details:
     \url{http://creativecommons.org/licenses/by/4.0/}.
   }

  Automatic lexical semantic change detection is a field of semantic analysis which aims to discern how the meanings of words change over time. As interest in the field has increased, a variety of different procedures, languages and corpora have been used, which leads to difficulty when attempting to  compare different sets of results. SemEval-2020 Task 1, ``Unsupervised Lexical Semantic Change Detection'' \cite{schlechtweg2020semeval}, is a task aimed at providing a single unified framework with which to compare approaches using a standardised dataset, in order to address the difficulty in attempting to  compare different sets of results which arises due to the variety in procedures, languages and corpora which have previously been used. The task involves determining whether a set of target words have changed meaning in two corpora, each of which corresponds to a different time period. Corpora are provided for German, English, Latin and Swedish. The task consists of two subtasks, one involving a binary classification of target words into words which have or have not changed meaning, the other involving a ranking of words according to degree of change. 
  
  This paper describes our submission to SemEval-2020 Task 1\footnote{Source code and data are published at \url{https://github.com/elerisarsfield/semeval}}. To produce this submission, we use an approach based on the Hierarchical Dirichlet Process (HDP) \cite{TehJorBea2004}, an extension of Latent Dirichlet Allocation (LDA) which allows the number of senses to be unbounded. The next section presents works related to methods for identifying novel word senses, which is followed by a description of our system in Section \ref{sec:methodology} and the results of the application of the system in Section \ref{sec:results}.  Section \ref{sec:conclusion} provides an overview of our conclusions and possible directions of future work. We also use our method on a corpus constructed from data from Twitter in  order to explore the possibility of detecting semantic change over a shorter period of time, with a focus on where that detection can be used for identifying slang. 

\section{Related Work}
\label{sec:related}
Approaches to lexical semantic change detection collect co-occurrence information on words, represented in most approaches by pointwise mutual information scores \cite{DBLP:journals/corr/abs-1811-06278}. This information may be used as the basis of an approach, as is used by \newcite{10.5555/1705415.1705429} and \newcite{gulordava-baroni-2011-distributional} to identify semantic change over time by using context vectors. Context vectors can also be used as part of a different representation. \newcite{DBLP:journals/corr/YaoSDRX17} use a PPMI matrix for each time period to learn dynamic embeddings, an approach to identifying lexical change which reduces the need to align vectors that arises when using the static embeddings which are traditionally used. \newcite{bamler2017dynamic} also use dynamic embeddings, presenting a probabilistic version of word2vec \newcite{mikolov2013efficient}, an architecture for static embeddings. Embeddings are also used by \newcite{asgari-etal-2020-unisent}, who use domain-specific embeddings to detect domain shifts. A  survey of the use of embeddings for semantic shifts detection is given in \newcite{DBLP:journals/corr/abs-1806-03537}.

Context vector and embeddings approaches to identifying change of meaning do not have the ability to recover the word senses. \newcite{Brody:2009:BWS:1609067.1609078} give an approach to the problem of word sense induction which is based on the Latent Dirichlet Allocation (LDA) model \cite{blei2003latent} of text generation. The LDA model is extended from parametric to nonparametric space by \newcite{zhai2013online} with an infinite vocabulary. \newcite{lau-etal-2012-word} instead use a Hierarchical Dirichlet Process (HDP) model \cite{TehJorBea2004} for the task of word sense induction. A model based on the HDP is similarly used by   \newcite{yao2011nonparametric}, for extending the flexibility of word sense induction models in order to adapt to varying degrees of polysemy. An HDP is also used by \newcite{cook-etal-2014-novel}, who apply word sense induction to dictionaries and newspapers. A similar approach is taken by \newcite{cook2013lexicographic}, in the context of updating dictionaries. \newcite{doi:10.1162/tacl_a_00081} use a Bayesian model for tracking gradual sense changes over time.

Graph-based models to finding lexical semantic change are proposed by \newcite{tahmasebi2011towards}, tracking language evolution by clustering word senses over various time periods, an approach echoed by  \newcite{DBLP:journals/corr/MitraMRBMG14}, studying word sense change in digitised books from 1520 to 2008. The most comprehensive survey of the field of lexical semantic change detection is presented by \newcite{DBLP:journals/corr/abs-1811-06278}. 

\section{System Overview}
\label{sec:methodology}

The data from SemEval-2020 Task 1 consists of two corpora for each language, each consisting of a number of short contexts of words. Though data are provided in a number of languages, only the data for English was used. Of each of the two corpora, one corpus corresponds to a reference corpus and the other to a focus corpus. The reference corpus is taken to represent standard usage, and the focus corpus contains newer texts.  Each corpus is partitioned into short pseudo-documents, which may be treated as documents. `Pseudo-document' and `document' will hereafter be used interchangeably. The creation of these pseudo-documents is under the assumption that each corpus takes the form of short contexts of a small number of sentences of text.  All stopwords and low frequency terms are removed.

\begin{table}[h]
  \centering
  \begin{tabular}{|l| l|}
    \hline    Notation&Description\\\hline
    $n_{jt}$&Number of words in document $j$ at table $t$\\\hline
    $n_{jt}^{-ji}$&Number of words in document $j$ at table $t$ except $x_{ji}$\\\hline
    $m_{\cdot k}$&Number of tables with topic  $k$ in the corpus\\\hline
    $m_{\cdot\cdot}$&Number of tables in the corpus\\\hline
    $f_k^{-x_{ji}}(x_{ji})$&Conditional density of $x_{ji}$ under mixture component $k$ given all items except $x_{ji}$\\\hline 
    $f_k^{-\bm{x}_{jt}}(\bm{x}_{jt})$&Conditional density of $\bm{x}_{jt}$ given all items associated with $k$ except $\bm{x}_{jt}$\\\hline 
  \end{tabular}
  \caption{Notation used in the Chinese Restaurant Franchise}
  \label{tab:not}
\end{table}
A Word Sense Induction (WSI) model  based on a Hierarchical Dirichlet Process (HDP), a clustering approach which allows clusters to be shared among the groups \cite{TehJorBea2004}, is used to model the senses for each word. Notation is given in Table \ref{tab:not}. An HDP is selected over a Latent Dirichlet Allocation (LDA) model as used in \newcite{Brody:2009:BWS:1609067.1609078} because the HDP allows the number of latent factors in the HDP to grow with the data.
\begin{figure}
  \centering
  \includegraphics[scale=0.5]{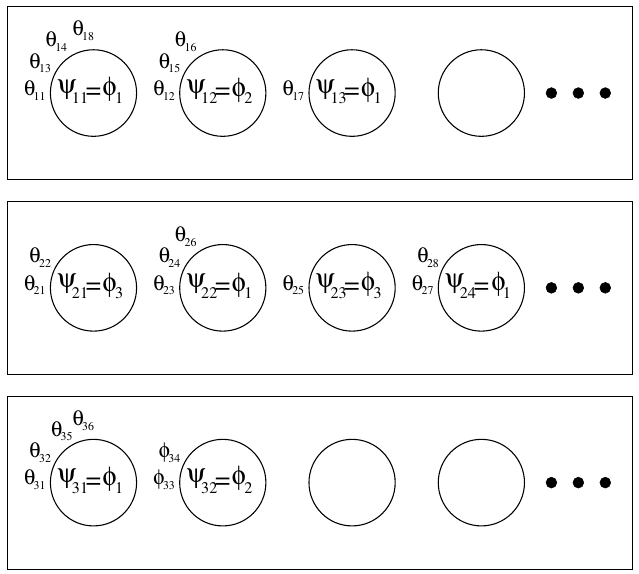}
  \caption{A depiction of a Chinese Restaurant Franchise \cite{TehJorBea2004}. Customers $\theta_i$ sit at tables $\psi_t$ which each order menu items $\phi_k$}
  \label{fig:restaurant}
\end{figure}

An initial random set of senses is induced modelled after the generative process of the HDP, which corresponds to a partition of words inside documents, a process based on the Chinese Restaurant Franchise (CRF)  representation of a two level HDP \cite{TehJorBea2004}, which partitions customers  at the group level and dishes at the top level. Each document $j$ may be considered a group. Its document level Chinese Restaurant Process (CRP) generates a table index $t_{ji}$ for each observation $i$ according to
\begin{equation}
  p(t_{ji}=t|t_{j1},\cdots,t_{j,i-1},\alpha)\propto
  \begin{cases}
    n_{jt}&\text{if }t\text{ previously used}\\
    \alpha&\text{if }t=t^{new}
  \end{cases}
\end{equation}
where $\alpha$ is a concentration parameter. This partitions the document $j$ into tables.

After all words are assigned to a table, the topic index $k_{jt}$ of all tables are generated by the corpus-level CRP. Each table is assigned a topic index according to
\begin{equation}
  p(k_{jt}|k_{11},\cdots,k_{j,t-1},\gamma)\propto
  \begin{cases}
    m_{\cdot k}&\text{if }k\text{ previously used}\\
    \gamma&\text{if }k=k^{new}
  \end{cases}
\end{equation}
where $\gamma$ is a concentration parameter for the base distribution. This assigns tables to topics, which completes the initial partition. Each observation $x_{ji}$ is associated with a table index $t_{ji}$, which is in turn associated with a topic index $k_{jt}$.  This topic index links the table to one of the corpus topics $\phi_k$. This partition is updated using the sampling scheme outlined in \newcite{TehJorBea2004}, which we summarise shortly.

Each word in each document is assigned to a per-document table $t_{ji}$, with each of these tables associated with a factor $k_{jt}$. The conditional distribution of $t_{ji}$ is calculated by combining the conditional prior distribution for $t_{ji}$ with the likelihood of generating $x_{ji}$. The prior probability of $t_{ji}$ taking a previously used value is proportional to $n_{jt}^{-ji}$. The probability it takes on a new value is proportional to the parameter $\alpha$. The likelihood due to $x_{ji}$  of $t_{ji}$ for a previously used $t$ is $f_k^{-x_{ji}}(x_{ji})$. For $t_{ji}=t^{new}$ the likelihood is
\begin{equation}
  p(x_{ji}|\bm{t}^{-ji},t_{ji}=t^{\text{new}},\bm{k})=\sum_{k=1}^{K}\frac{m_{\cdot k}}{m_{\cdot\cdot}+\gamma}f_k^{-x_{ji}}(x_{ji})+\frac{\gamma}{m_{\cdot\cdot}+\gamma}f_{k^{\text{new}}}^{-x_{ji}}(x_{ji})
\end{equation}
where $f_{k^{\text{new}}}^{-x_{ji}}(x_{ji})=\int f(x_{ji}|\phi)h(\phi)d\phi$ is the prior density of $x_{ji}$. From this, the conditional distribution of $t_{ji}$ can be formed as
\begin{equation}
  p(t_{ji}=t|\bm{t}^{-ji},\bm{k})\propto
  \begin{cases}
    n_{jt}^{-ji}f_{k_{jt}}^{-x_{ji}}(x_{ji})&\text{if }t\text{ previously used}\\
    \alpha p(x_{ji}|\bm{t}^{-ji},t_{ji}=t^{\text{new}},\bm{k})&\text{if }t=t^{\text{new}}
  \end{cases}
\end{equation}
This provides the distribution according to which the tables for words are sampled. The PPMI values in the co-occurrences matrix are used for estimation. If the sampled value is $t^{\text{new}}$, then a new table is created for the restaurant and the word assigned to it. A dish must then be selected for this table. A sample for $k_{jt^{\text{new}}}$ is obtained according to
  \begin{equation}
    p(k_{jt^{\text{new}}}=k|\bm{t},\bm{k}^{-jt^{\text{new}}})\propto
    \begin{cases}
      m_{\cdot k}f_k^{-x_{ji}}(x_{ji})&\text{if } k\text{ previously used}\\
      \gamma f_{k^{\text{new}}}^{-x_{ji}}(x_{ji})&\text{if }k=k^{\text{new}}
    \end{cases}
  \end{equation}
  This gives an assignment of each word to a cluster inside a document, and of each cluster to a corpus-level topic. Senses are formed from each topic by considering  the distribution of assigned words.  These topics represent the senses across both corpora, inferred on pooled instances of each word. Since the topics are jointly modelled, discovered topics are applicable to both corpora, which means that there is no need to reconcile these senses \cite{cook-etal-2014-novel}. Words with multiple senses are represented by their position in multiple topics. Once sampling has concluded, the word distributions over topics are calculated. Word distributions are calculated based on the topics assigned to each observed word $x_{ji}$ in each document to determine a distribution over topics separately for each word for the reference and focus instances. From this, the Jensen-Shannon distance for the target words between the distribution in the reference and focus corpus is calculated.

  Additionally, a method for the application to the Twitter corpus  is used based on the Novelty\textsubscript{Diff} method in \newcite{cook-etal-2014-novel}, a description of which follows.  The distribution of  words over  senses is categorised according to which instances originated in the reference corpus and which in the focus corpus.  For each sense $s$, the novelty score is calculated as
\begin{equation}
  \text{Novelty}_{\text{Diff}}(s)=p_f(s)-p_r(s)
\end{equation}
where $p_f(s)$ and $p_r(s)$ are the proportion of usages of a given word corresponding to sense $s$ in the focus and reference corpus, respectively. The score for each word is the maximum  for any of its induced senses. Once the novelty scores have been calculated for all words, the words with novel senses can be identified as those with the highest scores of Novelty\textsubscript{Diff}. This will be used when applying the model to Twitter data rather than the SemEval task in order to identify which words have gained senses, a narrower focus which is useful in the applied context of identifying slang.

\section{Results}
\label{sec:results}
 
 The model's performance on SemEval-2020 Task 1 is presented in  Table \ref{tab:semeval},  alongside results from the baselines for each subtask. The model was only run against the English data for each subtask, using the sample files for all other languages to form a complete submission. To reflect this, we record both the overall score and the score based on performance on the English component in Table \ref{tab:semeval}. The frequency difference baseline is the absolute difference in normalised target word frequencies in the corpora. The count vector baseline is the cosine difference between vector representations of words in the two corpora. The values for these baselines are provided by the task organisers\footnote{\url{https://competitions.codalab.org/competitions/20948\#results}}. The model was  run for 1 iteration. The window size was set to 2, the floor for determining low frequency terms was set to 1, and the concentration parameters $\alpha$ and $\gamma$ were both set to 1. To produce the submission, the Jensen-Shannon score for each target is calculated. This  provides the submission for Subtask 2. For Subtask 1, lemmas are determined to have changed meaning if their Jensen-Shannon distance is above a threshold value of 0.6. 
\begin{table}[h]
  \centering
  \begin{tabular}{|c|c|c|c|c|c|}\hline
    Subtask&Score (Overall)&Score (English)&Frequency Difference&Count Vector&Rank\\\hline
    1&0.526&0.568&0.432&0.595&19\\\hline
    2&0.100&0.105&-0.217&0.022&18\\\hline
  \end{tabular}
  \caption{Results for SemEval 2020 Task 1 for English data. Baselines scores are taken from the English component only.}
  \label{tab:semeval}
\end{table}
The results from the SemEval competition (see Table \ref{tab:semeval} for complete results), for which we only produced a submission for the English component, led to a placement, when  considering performance on the English data only, of  7\textsuperscript{th} of 9\textsuperscript{th} for the classification subtask and 16\textsuperscript{th} of 21\textsuperscript{st} for the ranking subtask, and were more accurate than both of the baselines for the ranking subtask, and for one of the baselines for the classification subtask, indicating the model works as expected to a reasonable degree of accuracy. The lacklustre  performance on the first subtask as compared with the second may have been influenced by the choice of threshold parameter, which was never altered and chosen with little empirical evidence, which resulted in a threshold parameter which was too low and as such was unable to appropriately identify instances of meaning change. Experimenting  with this value, as well as those of other hyperparameters, would likely have improved performance.
\begin{table}[ht]
  \centering      
  \begin{tabular}{|l| p{13cm}|}
\hline    \textbf{Word}&\textbf{Novel Definition}\\\hline
    ill&cool, tight, sweet\\\hline
    hater&a person that simply cannot be happy for another person's success. So rather than be happy they make a point of exposing a flaw in that person. \\\hline
        like&a meaningless word teenagers insert liberally into both colloquial and formal speech in order to maintain a steady stream of words\\\hline
        roll&used to describe the effects of Ecstasy\\\hline
        safe&a cool person;to signify agreement;to signify something is good\\\hline
        checked&the process by which someone puts another individual in their place verbally or physically either in a joking manner or a serious beatdown.\\\hline
        bars&(1) sentences in lyrical hiphop songs (2) slang name for a 2mg Xanax tablet\\\hline 
  \end{tabular}
  \caption{Identified words and definitions}
  \label{tab:words}
\end{table}

In addition to the SemEval competition, the same model was also used to experiment with data from Twitter with the goal of determining whether the same procedure can be used for sense change over a shorter timeframe, with a particular focus on identifying slang through words which had novel senses. The corpus was constructed from a set of 500,000 tweets, spanning a period of 15 years. Using the approach detailed in Section \ref{sec:methodology} with the Twitter data produced a set of 7 slang words. These results are the words identified as having the most difference in the senses between the two corpora according to the Novelty\textsubscript{Diff} metric. The model was  run for 1 iteration against the Twitter  corpora  in order to identify the lemmas determined as having most differed between the older and more recent corpus, and therefore most likely to have gained a novel sense. The 25 words with the highest value of $\text{Novelty}_{\text{Diff}}$ were extracted. The online slang dictionary Urban Dictionary\footnote{\url{https://www.urbandictionary.com/}} was used to verify whether or not each of these words had a meaning aside from its standard usage or was generated erroneously. Where an identified word was a genuine instance of slang, it is listed alongside this sense as defined by Urban Dictionary in Table \ref{tab:words}. Of the identified words, 7 were able to be verified.  The full set of words which were identified is available on GitHub.

\section{Conclusions and Future Work}
\label{sec:conclusion}
This paper used Bayesian word sense induction methods as the basis for identifying lexical semantic change, and produced a model for determining sense differences which was then evaluated on Task 1 in the SemEval-2020 workshop, which confirmed the method is appropriate to the task and works as expected. The same approach was also used with data from Twitter with the goal of identifying slang, which produced a set of candidate words. Whilst some of these were genuine instances of slang, a significant amount were erroneously identified, or amounted to mere lexical innovation rather than any true semantic change. 

The system was only tested against the English data for each subtask. Future work would involve extending the system into all of the competition's languages.
\nocite{article}
\bibliographystyle{coling}
\bibliography{semeval2020}

\begin{thebibliography}{}

\bibitem[\protect\citename{Asgari \bgroup et al.\egroup
  }2020]{asgari-etal-2020-unisent}
Ehsaneddin Asgari, Fabienne Braune, Benjamin Roth, Christoph Ringlstetter, and
  Mohammad Mofrad.
\newblock 2020.
\newblock {U}ni{S}ent: Universal adaptable sentiment lexica for 1000+
  languages.
\newblock In {\em Proceedings of The 12th Language Resources and Evaluation
  Conference}, pages 4113--4120, Marseille, France, May. European Language
  Resources Association.

\bibitem[\protect\citename{Bamler and Mandt}2017]{bamler2017dynamic}
Robert Bamler and Stephan Mandt.
\newblock 2017.
\newblock Dynamic word embeddings.
\newblock In {\em Proceedings of the 34th International Conference on Machine
  Learning-Volume 70}, pages 380--389. JMLR. org.

\bibitem[\protect\citename{Blei \bgroup et al.\egroup }2003]{blei2003latent}
David~M Blei, Andrew~Y Ng, and Michael~I Jordan.
\newblock 2003.
\newblock Latent dirichlet allocation.
\newblock {\em Journal of machine Learning research}, 3(Jan):993--1022.

\bibitem[\protect\citename{Brody and
  Lapata}2009]{Brody:2009:BWS:1609067.1609078}
Samuel Brody and Mirella Lapata.
\newblock 2009.
\newblock Bayesian word sense induction.
\newblock In {\em Proceedings of the 12th Conference of the European Chapter of
  the Association for Computational Linguistics}, EACL '09, pages 103--111,
  Stroudsburg, PA, USA. Association for Computational Linguistics.

\bibitem[\protect\citename{Cook \bgroup et al.\egroup
  }2013]{cook2013lexicographic}
Paul Cook, Jey~Han Lau, Michael Rundell, Diana McCarthy, and Timothy Baldwin.
\newblock 2013.
\newblock A lexicographic appraisal of an automatic approach for detecting new
  word senses.
\newblock {\em Proceedings of eLex}, pages 49--65.

\bibitem[\protect\citename{Cook \bgroup et al.\egroup
  }2014]{cook-etal-2014-novel}
Paul Cook, Jey~Han Lau, Diana McCarthy, and Timothy Baldwin.
\newblock 2014.
\newblock Novel word-sense identification.
\newblock In {\em Proceedings of {COLING} 2014, the 25th International
  Conference on Computational Linguistics: Technical Papers}, pages 1624--1635,
  Dublin, Ireland, August. Dublin City University and Association for
  Computational Linguistics.

\bibitem[\protect\citename{Frermann and Lapata}2016]{doi:10.1162/tacl_a_00081}
Lea Frermann and Mirella Lapata.
\newblock 2016.
\newblock A bayesian model of diachronic meaning change.
\newblock {\em Transactions of the Association for Computational Linguistics},
  4:31--45.

\bibitem[\protect\citename{Gulordava and
  Baroni}2011]{gulordava-baroni-2011-distributional}
Kristina Gulordava and Marco Baroni.
\newblock 2011.
\newblock A distributional similarity approach to the detection of semantic
  change in the {G}oogle books ngram corpus.
\newblock In {\em Proceedings of the {GEMS} 2011 Workshop on {GE}ometrical
  Models of Natural Language Semantics}, pages 67--71, Edinburgh, UK, 7.
  Association for Computational Linguistics.

\bibitem[\protect\citename{Kutuzov \bgroup et al.\egroup
  }2018]{DBLP:journals/corr/abs-1806-03537}
Andrey Kutuzov, Lilja {\O}vrelid, Terrence Szymanski, and Erik Velldal.
\newblock 2018.
\newblock Diachronic word embeddings and semantic shifts: a survey.
\newblock {\em CoRR}, abs/1806.03537.

\bibitem[\protect\citename{Lau \bgroup et al.\egroup }2012]{lau-etal-2012-word}
Jey~Han Lau, Paul Cook, Diana McCarthy, David Newman, and Timothy Baldwin.
\newblock 2012.
\newblock Word sense induction for novel sense detection.
\newblock In {\em Proceedings of the 13th Conference of the {E}uropean Chapter
  of the Association for Computational Linguistics}, pages 591--601, Avignon,
  France, April. Association for Computational Linguistics.

\bibitem[\protect\citename{Mikolov \bgroup et al.\egroup
  }2013]{mikolov2013efficient}
Tomas Mikolov, Kai Chen, Greg Corrado, and Jeffrey Dean.
\newblock 2013.
\newblock Efficient estimation of word representations in vector space.

\bibitem[\protect\citename{Mitra \bgroup et al.\egroup
  }2014]{DBLP:journals/corr/MitraMRBMG14}
Sunny Mitra, Ritwik Mitra, Martin Riedl, Chris Biemann, Animesh Mukherjee, and
  Pawan Goyal.
\newblock 2014.
\newblock That's sick dude!: Automatic identification of word sense change
  across different timescales.
\newblock {\em CoRR}, abs/1405.4392.

\bibitem[\protect\citename{Sagi \bgroup et al.\egroup
  }2009]{10.5555/1705415.1705429}
Eyal Sagi, Stefan Kaufmann, and Brady Clark.
\newblock 2009.
\newblock Semantic density analysis: Comparing word meaning across time and
  phonetic space.
\newblock In {\em Proceedings of the Workshop on Geometrical Models of Natural
  Language Semantics}, GEMS ’09, page 104–111, USA. Association for
  Computational Linguistics.

\bibitem[\protect\citename{Schlechtweg \bgroup et al.\egroup
  }2020]{schlechtweg2020semeval}
Dominik Schlechtweg, Barbara McGillivray, Simon Hengchen, Haim Dubossarsky, and
  Nina Tahmasebi.
\newblock 2020.
\newblock {S}em{E}val-2020 {T}ask 1: {U}nsupervised {L}exical {S}emantic
  {C}hange {D}etection.
\newblock In {\em To appear in Proceedings of the 14th International Workshop
  on Semantic Evaluation}, Barcelona, Spain. Association for Computational
  Linguistics.

\bibitem[\protect\citename{Tahmasebi \bgroup et al.\egroup
  }2011]{tahmasebi2011towards}
Nina Tahmasebi, Thomas Risse, and Stefan Dietze.
\newblock 2011.
\newblock Towards automatic language evolution tracking, a study on word sense
  tracking.
\newblock In {\em Joint Workshop on Knowledge Evolution and Ontology Dynamics}.

\bibitem[\protect\citename{Tahmasebi \bgroup et al.\egroup
  }2018]{DBLP:journals/corr/abs-1811-06278}
Nina Tahmasebi, Lars Borin, and Adam Jatowt.
\newblock 2018.
\newblock Survey of computational approaches to lexical semantic change.
\newblock {\em CoRR}, abs/1811.06278.

\bibitem[\protect\citename{Teh \bgroup et al.\egroup }2004]{TehJorBea2004}
Y.~W. Teh, M.~I. Jordan, M.~J. Beal, and D.~M. Blei.
\newblock 2004.
\newblock Hierarchical {D}irichlet processes.
\newblock Technical Report 653, Department of Statistics, University of
  California at Berkeley.

\bibitem[\protect\citename{Wang and Blei}2012]{article}
Chong Wang and David Blei.
\newblock 2012.
\newblock A split-merge mcmc algorithm for the hierarchical dirichlet process.
\newblock 01.

\bibitem[\protect\citename{Yao and Van~Durme}2011]{yao2011nonparametric}
Xuchen Yao and Benjamin Van~Durme.
\newblock 2011.
\newblock Nonparametric bayesian word sense induction.
\newblock In {\em Proceedings of TextGraphs-6: Graph-based Methods for Natural
  Language Processing}, pages 10--14. Association for Computational
  Linguistics.

\bibitem[\protect\citename{Yao \bgroup et al.\egroup
  }2017]{DBLP:journals/corr/YaoSDRX17}
Zijun Yao, Yifan Sun, Weicong Ding, Nikhil Rao, and Hui Xiong.
\newblock 2017.
\newblock Discovery of evolving semantics through dynamic word embedding
  learning.
\newblock {\em CoRR}, abs/1703.00607.

\bibitem[\protect\citename{Zhai and Boyd-Graber}2013]{zhai2013online}
Ke~Zhai and Jordan Boyd-Graber.
\newblock 2013.
\newblock Online latent dirichlet allocation with infinite vocabulary.
\newblock In {\em International Conference on Machine Learning}, pages
  561--569.

\end{thebibliography}
\end{document}